\documentclass[10pt,twocolumn,letterpaper]{article}
\usepackage{wacv}
\usepackage{times}
\usepackage{epsfig}
\usepackage{graphicx}
\usepackage{amsmath}
\usepackage{amssymb}
\usepackage{subfig}
\usepackage{graphicx}
\usepackage{amsmath}
\usepackage{amssymb}
\usepackage{mathtools} 
\usepackage{algpseudocode}	
\usepackage{comment, url}
\usepackage{bm}
\usepackage[ruled, vlined, linesnumbered]{algorithm2e}

\usepackage[hidelinks]{hyperref}

\wacvfinalcopy 

\def\wacvPaperID{****} 

\ifwacvfinal\fi
\begin{document}

\title{{\it E}nKCF: Ensemble of Kernelized Correlation Filters for
  High-Speed Object Tracking}

\author{Burak Uzkent \\
{\tt\small bxu2522@rit.edu}
\and
YoungWoo Seo \\
{\tt\small youngwoo.blank.seo@gmail.com}
}

\maketitle
\ifwacvfinal\thispagestyle{empty}\fi

\begin{abstract}
Computer vision technologies are very attractive for practical
applications running on embedded systems. For such an application, it
is desirable for the deployed algorithms to run in high-speed and
require no offline training. To develop a single-target tracking
algorithm with these properties, we propose an ensemble of the
kernelized correlation filters (KCF), we call it {\it E}nKCF. A
committee of KCFs is specifically designed to address the variations
in scale and translation of moving objects. To guarantee a high-speed
run-time performance, we deploy each of KCFs in turn, instead of
applying multiple KCFs to each frame. To reduce any potential drifts
between individual KCFs' transition, we developed a particle
filter. Experimental results showed that the performance of ours is,
on average, 70.10\% for precision at 20 pixels, 53.00\% for success
rate for the OTB100 data, and 54.50\% and 40.2\% for the UAV123
data. Experimental results showed that our method is better than other
high-speed trackers over 5\% on precision on 20 pixels and 10-20\% on
AUC on average. Moreover, our implementation ran at 340 fps for the
OTB100 and at 416 fps for the UAV123 dataset that is faster than DCF
(292 fps) for the OTB100 and KCF (292 fps) for the UAV123. To increase
flexibility of the proposed {\it E}nKCF running on various platforms,
we also explored different levels of deep convolutional features.
\end{abstract}

\section{Introduction}
\label{introduction}
A recent advancement of air/ground/water unmanned vehicle technologies
has increased interests on deploying intelligent algorithms to
existing embedded and mobile platforms. Among those technologies,
computer vision algorithms are getting more attentions primarily
because payloads of those mobile platforms are limited to carry any
other sensors than a monocular camera. Instead of just manually being
flew for video recording, an unmanned air vehicle (UAV) equipped with
an object or feature following function would make it more useful for
the application of monitoring/surveillance/surveying on private
properties/wild-life/crop, video recording on sports/movies/events,
many others. To this end, in this paper, we propose a single-target
tracking algorithm that does not require offline training and can run
in high-speed. Specifically, we would like to have our algorithm 1)
learn the appearance model of a target on the fly and 2) run on a
typical desktop as fast as $300$-$450$ fps.

\begin{figure*}[!h]
\centering
\includegraphics[width=\textwidth]{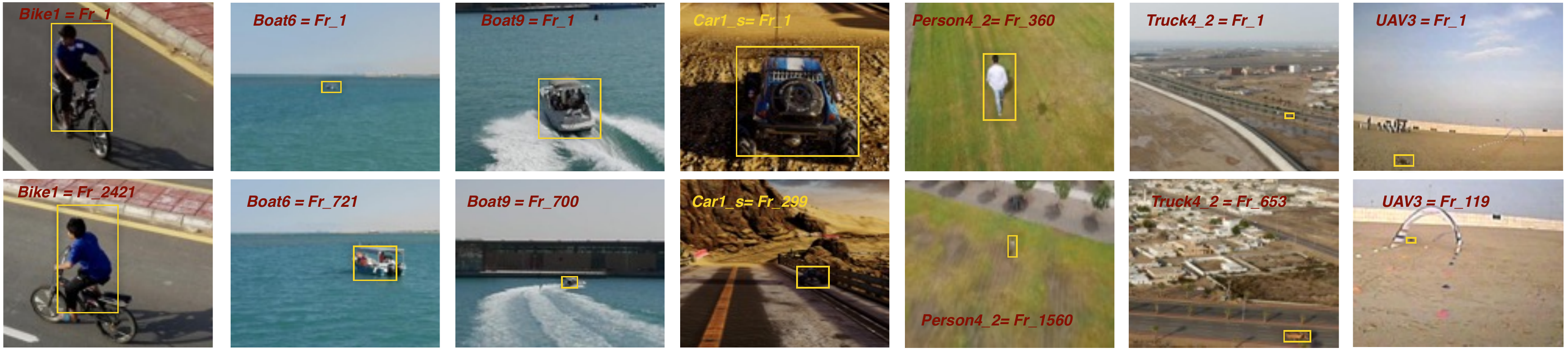}
\caption{Examples of some tracking results (yellow rectangles) by the
  proposed method on the ``UAV123'' dataset. The ``UAV123'' dataset is
  challenging for object tracking as the scale and translation of a
  target can be drastically changed in a few frames.}
\label{ResultsIntroduction}
\end{figure*}

\begin{figure*}[!h]
\centering
\begin{tabular}{cc}
\includegraphics[width=14.00cm]{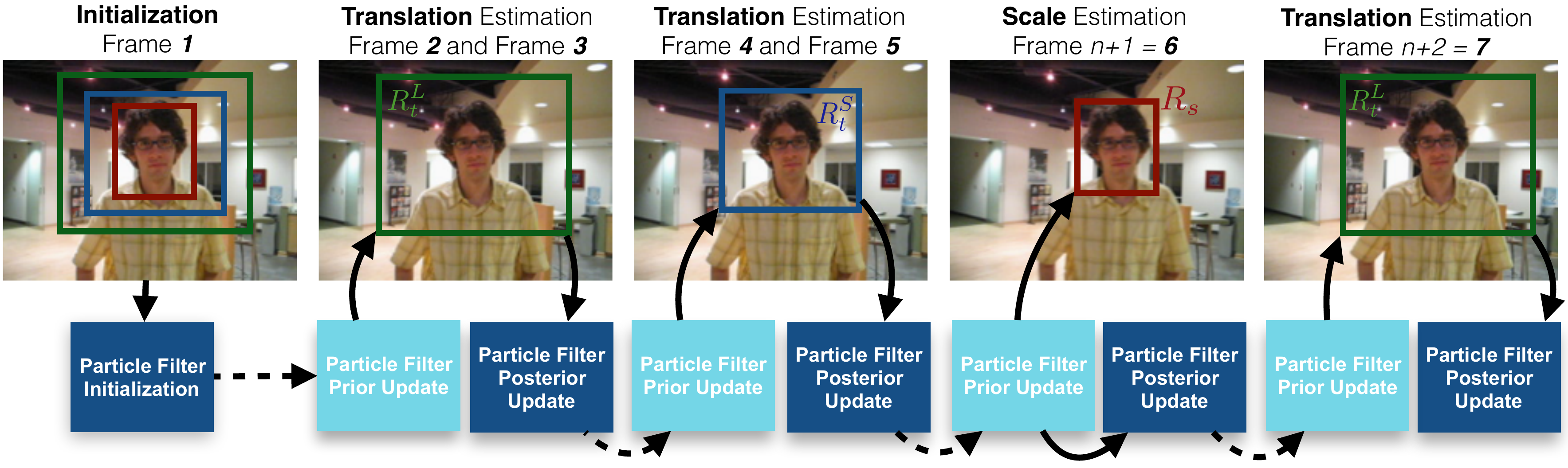}\\
\end{tabular}
\caption{The workflow of the {\it E}nKCF. The first frame is used to
  initialize the tracking algorithm and a particle filter. For the
  next six frames, each of three KCF is deployed in turn to estimate
  the translation and scale change of a target. Afterward, the order
  of deploying three KCFs is repeating.}
\label{Workflows}
\end{figure*}

One of the dominant frameworks for online object tracking is the
correlation filter that essentially solves a single-target tracking
problem as a regression problem in the frequency domain. This
framework assumes that a target location is given at the beginning
like any other online tracking algorithms
\cite{smeulders2014survey}. Given this positive example for the
regression problem, a set of negative examples is collected around the
initial target bounding box and represented as a form of the circulant
matrix \cite{henriques2015high}. One can optimally solve this
regression problem using a ridge regression in a closed form. However,
this solution has to deal with expensive matrix operations
$\mathcal{O}(n^{3})$. The correlation filter offers a less complex
solution, $\mathcal{O}(n\log n)$ over element-wise multiplication in a
frequency domain \cite{bolme2010visual,henriques2015high}. Thank to
this reformulation, an object tracking pipeline based on the
correlation filter can run very efficiently and be even easily
implemented. In fact, an extension of a linear correlation filter, the
kernelized correlation filter with multi-channel features
\cite{henriques2015high} showed impressive object tracking results and
outperformed other state-of-the-art, online tracking algorithms in
terms of run-time and tracking accuracy. However, a vanilla form of
such an online tracker is prone to drift, and fails to track a target
over a long period of time \cite{henriques2015high}. This is primarily
due to the dilemma of stability-plasticity in updating appearance
model, where the appearance model will be overfitted to only the
images used to train, unless a compromise on the frequency of updating
the model is carefully implemented \cite{santner2010prost}. For
example, one may handle a target's scale variation by just
concatenating multiple correlation filters including KCF and running
them on each frame. Alternatively one could think of scanning the
region of interest (ROI) with a list of templates in predefined scale
ratios to find the target in appropriate scale
\cite{henriques2015high,tang2015multi,ma2015long,bibi2015multi,li2014scale}. However,
these approaches would drastically reduce run-time performance because
multiple KCFs run on each frame. 

Another way of handling the scale change for the correlation filter
based approach is to estimate a correct scale at a location where a
target highly likely appears \cite{zhang2014fast} -- estimate a
target's translation first and then estimate correct scale. For
example, Zhang and his colleagues used the MOSSE tracker
\cite{bolme2010visual} to estimate the target's translation. And then
they attempted to update the scale of the target by further analyzing
image sub-regions in high confidence. Their method is based on an
assumption that the scale of a target would not change drastically over two
consecutive frames. Similarly, Ma and his colleagues used two KCFs to
learn the translation and scale of a target separately
\cite{ma2015long}. In particular, a KCF is used to learn the
translation of the target and its background. Given this, another KCF
is used to learn the target area to estimate the new scale of the
target. However, because of running more than a KCF on each frame,
this method degrades its run-time performance (i.e., $\leq 50
fps$). Our method is motivated by this idea -- the idea of deploying
multiple KCFs to address the issues of single target tracking: scale
and translation, but in a more efficient way. To maximize run-time
performance and accuracy, instead of running them all together on
every frame, we deploy three KCFs in turn:
\textit{target}+\textit{small background} translation filter
($R_{t}^{S}$), \textit{target-only} scale filter ($R_{s}$) and
\textit{target}+\textit{large background} translation filter
($R_{t}^{L}$). By doing so, our method aims at addressing scale
change and estimating target's motion efficiently while maintaining or
increasing run-time performance. Figure \ref{Workflows} illustrates
the workflow of the {\it E}nKCF.

The contribution of this paper is a novel, single-target tracking
algorithm running in a very high-speed ($\geq300$ fps). In particular,
to effectively address the changes in scale and translation of a
target, we extended the KCF and deployed each of three KCFs in
turn. Because of such a deployment strategy, the run-time performance
of the proposed algorithm maintains high without sacrificing the
accuracy of tracking. To reduce any potential drifts while switching
KCFs, we developed a particle filter. Additionally, to increase the
flexibility of the proposed algorithm's usage, we explore deep
convolutional features with varying levels of abstraction.

\section{{\it E}nKCF: Ensemble of Kernelized Correlation Filters}
We propose a novel way of utilizing an ensemble of the KCFs
\cite{henriques2015high} to effectively handle scale variations and
dynamic maneuvers of a target. To maintain or improve the run-time
performance of the original KCF (e.g., $\ge$ 300), we deploy three
KCFs in turn, instead of applying them on the same image frame
together. Each of these KCFs is designed to address the challenges of
single-target tracking -- variance in scale and translation. By
deploying each of three KCFs in turn, our algorithm will improve the
performance of the original KCF while preserving the original KCF's
high run-time performance and small-footprint.

\begin{algorithm*}[h]
\small
\DontPrintSemicolon
\KwIn{ Initial bounding box $(x_{0},y_{0},s_{0})$, frame counter $fc$, complete cycle of scale filter $n = 5$,}
\KwOut{	\uIf{$fc\:\%\:n=0$ (\textbf{Condition 1})}
			{
				Estimated Target State $(x_{t},y_{t},s_{t})$,
				Scale filter (\textit{target-only}) model $R_{s}$
			}
			 \ElseIf{$fc\:\%\:n>0\:\:and\:\:fc\:\%\:n\leq n/2$ (\textbf{Condition 2})}
			 {
				Estimated Target State $(x_{t},y_{t},s_{t} = s_{t-1})$,
			  	Large Area Translation Filter model $R_{t}^{L}$
			}			 
			 \Else (\textit{\textbf{Condition 3}})
			{			 
				 Estimated Target State $(x_{t},y_{t},s_{t} = s_{t-1})$,
				Small Area Translation Filter model $R_{t}^{S}$			
			}	
		}
\SetKwBlock{Begin}{function}{end function}
\Begin($\text{track($x_{t-1},y_{t-1},s_{t-1}$) }$)
{
 	// \textit{Translation Estimation - Particle Filter} \\
	Transit Particle Filter to the frame $t$ and compute the mean of prior pdf $(x_{t},y_{t},s_{t-1})$ \\
     // \textit{Translation Estimation - Correlation Filter} \\
	Crop the ROI for the $R_{t}^{L}$ (Condition 2), or $R_{t}^{S}$ (Condition 3) given $(x_{t},y_{t})$ and estimate new position as $(x_{t},y_{t})$ = $\mathbf{\textit{max}(y_{R_{t}})}$\\
	// \textit{Scale Estimation - Correlation Filter}\\
	Scale pool for $R_{s}$ : $\mathbf{S} = \lbrace1.05,1.0,1/1.05\rbrace$,\\
	Crop the ROI for the $R_{s}^{i}$ (Condition 1) and estimate scale factor, $\mathbf{\alpha}$ = $\underset{i\in S}{\textit{argmax}}(PSR(\boldsymbol{y_{R_{s}^{i}}}))$, and new scale $s_{t} = \alpha * s_{t-1}$,\\ 
	// \textit{Update Translation - Particle Filter}\\
	Do Importance Re-sampling (if necessary) and compute the 
	mean of posterior pdf $(x_{t},y_{t})$\\
	// \textit{Model Update}\\
	Update $R_{t}^{S}$ (Condition 3),\\
	Update $R_{t}^{L}$ (Condition 2),\\
	\uIf{$PSR(\boldsymbol{y_{R_{s}}}) \geq T_{R_{s}}$}{
		Update $R_{s}$ (Condition 1)}
  \label{endfor}
  \Return{($x_{t},y_{t},s_{t}$)}
}
\caption{{\it E}nKCF Tracking Algorithm}\label{alg:MKCF}
\end{algorithm*}

The proposed algorithm, {\it E}nKCF, learns three KCFs in turn: The
first filter, $R_{t}^{S}$, focuses on learning the target area and its
background for addressing a marginal translation by a target, the
second filter, $R_{s}$, focuses entirely on learning the target's
scale, and the last filter, $R_{t}^{L}$, focuses on the target area
and its background bigger than that of the first filter,
$R_{t}^{S}$. We set up {\it E}nKCF in this way so that a correlation
filter for learning a target's translation could include the
background of the target to better discriminate it from its
surrounding, and another correlation filter could focus on the target
itself to estimate the right scale of the target. In particular, a
transition filter with larger padding size, $R_{t}^{L}$ will enable
{\it E}nKCF to recover from potential drifts after a scale filter is
applied to the input images. On the other hand, another translation
filter with smaller padding size, $R_{t}^{S}$, will help {\it E}nKCF
to better localize the position of the target after the large ROI translation
filter is applied.

\begin{figure*}[!h]
\centering
\includegraphics[width=1\textwidth]{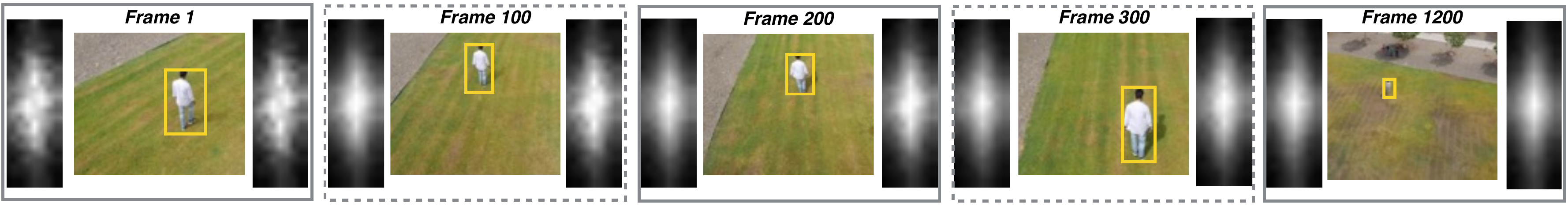}
\caption{These examples show that there is a marginal difference
  between the scale filters learned at every frame and the one learned
  at every 5 frames. The left, sub-figures show the scale filters
  trained at every frame and those at the right show the scale filters
  trained at every 5 frames.}
\label{fig:Filters_Comparison}
\end{figure*}

Our approach is similar to that of \cite{ma2015long} which
sequentially applied more than one KCF to every frame, but different
in that we operate multiple KCFs in an alternating manner. It
intuitively makes sense to alternatively use multiple KCFs with
different purposes because the appearance of a target does not change
drastically over consecutive images. In other words, for most cases,
learning a correlation filter over consecutive frames would not be
substantially different from the one updated with smaller
frequency. Figure \ref{fig:Filters_Comparison} shows examples
supporting this observation.

The algorithm \ref{alg:MKCF} shows the pseudo-code of {\it E}nKCF.
The order of running these three KCFs is important because each of
these filters aims at addressing different challenges. A translation
filter, $R_{t}^{L}$, is applied to the $i$th and $i$+$1$th image
frames, another translation filter, $R_{t}^{S}$, is applied to the
$i$+$2$th and $i$+$3$th image frames, and then the scale filter,
$R_{s}$, is applied to the $i$+$4$th image. This order repeats until
the last image is presented. Note that the translation filter,
$R_{t}^{L}$, is intended to run right after the scale filter, $R_s$,
runs which is applied at every other $i$+$4$ frames. We run these
filters in this way because we want to minimize any drifts that are
likely to happen running only $R_{s}$. In addition, the filter,
$R_{t}^{S}$, is applied to every other two frames before $R_{s}$ and
right after two consecutive frames running $R_{t}^{L}$. By repeating
this order of learning three KCFs, we can integrate more
discriminative shape features that cannot be learned just by
$R_{t}^{L}$. The filter, $R_{t}^{L}$, uses shape and color information
together to recover from any potential drifts -- drifts could happen
due to only scale filter operation in certain frames.

In summary, the scale filter, $R_{s}$, is designed to directly handle
the scale change of a target and provides the translation filters,
$R_{t}^{L}$ and $R_{t}^{S}$, with more accurate ROIs. On the other
hand, a translation filter, $R_{t}^{L}$, is devised to look into a
larger search area to estimate the target's translation and recover
from any potential drifts. Another translation filter, $R_{t}^{S}$, is
designed to address the drawback of $R_{t}^{L}$ that it may learn
noisy shape features due to its relatively larger search area. In what
follows, we will detail the main idea behind the KCF.

\textbf{Kernelized Correlation Filter} The Kernelized Correlation
Filter is a well-known single target tracking method and its workflow
has been detailed in other papers
\cite{henriques2012exploiting,henriques2015high}. This section briefly
goes over the parts of the KCF relevant to this study. Its
computational efficiency is derived from the correlation filter
framework representing training examples as a circulant matrix. The
fact that a circulant matrix can be diagonalized by Discrete Fourier
transform is the key to reduce the complexity of any tracking method
based on correlation filter. The off-diagonal elements become zero
whereas the diagonal elements represent the eigenvalues of the
circulant matrix. The eigenvalues are equal to the DFT transformation
of the base sample ($\boldsymbol{x}$) elements. The Kernelized
Correlation Filter, in particular, applies a kernel to
$\boldsymbol{x}$ to transform to a more discriminative domain. The
circulant matrix is then formed by applying cyclic shifts on the
kernelized $\boldsymbol{x}$. Such kernelization operation maintains
$\mathcal{O}$($nlog(n)$) complexity unlike other kernel algorithms
leading to $\mathcal{O}$($n^{2}$) or even higher complexities.

The KCF solves essentially the problem of a regression in the form of
the regularization (ridge regression):
\begin{equation}
E_h = \min_{\boldsymbol{h}}\frac{1}{2}||\boldsymbol{y}-\sum_{c=1}^{C}\boldsymbol{h_{c}}*\boldsymbol{x_{c}}||^{2} + \frac{\lambda}{2}\sum_{c=1}^{C}||\boldsymbol{h_{c}}||^{2}
\label{eq:Closedform_RidgeReg}
\end{equation}
where we seek for $\boldsymbol{h}$ that minimizes $E$ given the desired
continuous response, $\boldsymbol{y}$, and the training template
$\boldsymbol{x}$.  The parameter $c$ enables one to integrate features in
the form of multiple channels, such as HoG and color, in this setup
\cite{henriques2015high,galoogahi2013multi}.  A closed-form solution
for Equation \ref{eq:Closedform_RidgeReg} exists. To simplify the
closed-form solution, an element-wise multiplication in frequency
domain was proposed to learn the frequency domain correspondence of
$\boldsymbol{h}$, $\boldsymbol{\hat{w}}$:
\begin{equation}
\boldsymbol{\hat{w}} = \boldsymbol{\hat{x}^{*}}*\boldsymbol{\hat{y}}(\boldsymbol{\hat{x}^{*}}*\boldsymbol{\hat{x}}+\lambda)^{-1}.
\label{eq:DiagonalizedPrimalSolution}
\end{equation}
Where $*$ is an element-wise multiplication. A non-linear version of
$\boldsymbol{\hat{w}}$ and $\boldsymbol{\hat{\alpha}}$ are proposed to
increase robustness to any geometric and photometric variations
\cite{henriques2015high}. In particular, the diagonalized non-linear
Fourier domain dual form solution is expressed as
\begin{equation}
\boldsymbol{\hat{\alpha}} = \boldsymbol{\hat{y}}(\boldsymbol{\hat{k}^{xx^{'}}}+\lambda)^{-1}
\label{eq:FourierDualDomainSolution}
\end{equation}
where $\lambda$ represents the regularization weight whereas $\boldsymbol{\hat{k}^{xx^{'}}}$ denotes
 the first row of the kernel matrix $\boldsymbol{K}$ known as \textit{gram matrix} and is expressed as
\begin{equation}
\boldsymbol{k^{xx^{'}}} = exp(-\dfrac{1}{\alpha^{2}}(||\boldsymbol{x}||^{2}+||\boldsymbol{x}^{'}||^{2}-2F^{-1}(\sum^{C}_{c}\boldsymbol{\hat{x}_{c}}^{*}\odot \boldsymbol{\hat{x}_{c}}^{'}))).
\label{eq:GaussianCorrelationSingleChannel}
\end{equation}
An early version based on this formulation used grayscale feature
($C=1$) to learn the solution vector $\boldsymbol{w}$ and integrating
multi-channel features such as HoG and Color showed improved accuracy
\cite{henriques2015high,galoogahi2013multi,tang2015multi,ma2015long,bibi2015multi}.
In the detection phase, the learned correlation filter is correlated
with the first row of the gram matrix, $\boldsymbol{k^{xz^{'}}}$, which
contains the similarity values between the learned feature template
$\boldsymbol{x}$ and the new test template $\boldsymbol{z}$. This can be
formulated as
\begin{equation}
\boldsymbol{r(z)} = F^{-1}(\boldsymbol{\hat{k}^{xz^{'}}} \odot \boldsymbol{\hat{\alpha}})
\end{equation}
where $\boldsymbol{r}$ denotes the correlation response at all cyclic shifts of the
first row of the kernel matrix.

As a common practice, to integrate further temporal information into tracker, we
update the correlation filter and the appearance model as below.
\begin{align}
\boldsymbol{\hat{\alpha}_{t}} = (1-\beta) \boldsymbol{\hat{\alpha}_{t-1}} + \beta \boldsymbol{\hat{\alpha}_{t}} \\
\boldsymbol{\hat{x}_{t}} = (1-\beta) \boldsymbol{\hat{x}_{t-1}} + \beta \boldsymbol{\hat{x}_{t}}
\end{align}
where $\beta$ represents the learning rate tuned to a small value in practice.

In the following sections, we will detail two different types of
features used in {\it E}nKCF: hand-crafted features (fHoG +
color-naming) and deep convolutional features.

\begin{figure}[!h]
\centering
\includegraphics[width=0.47\textwidth]{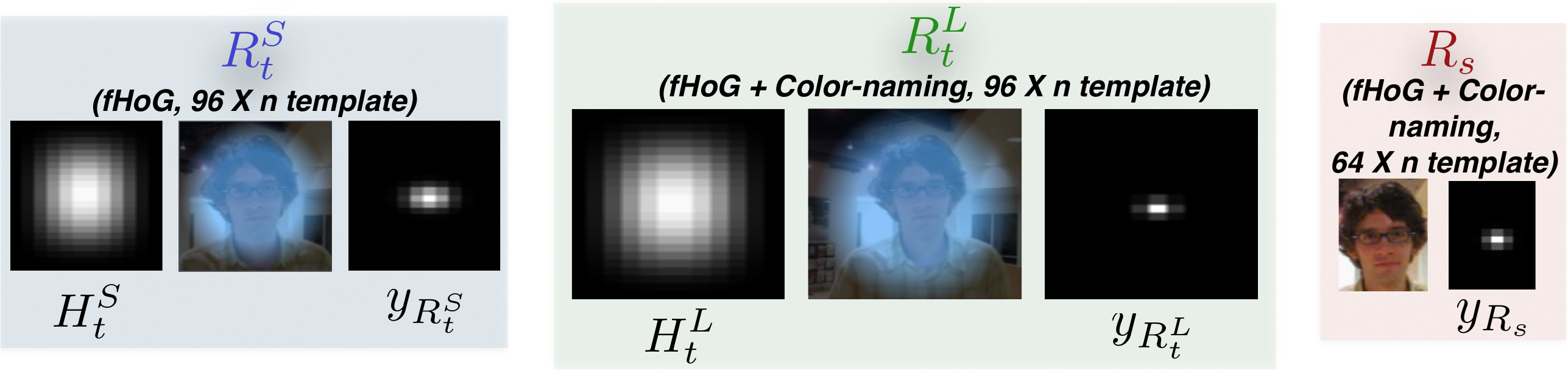}
   \\[-3ex]
\subfloat[Small Area Translation Filter]{\hspace{.30\linewidth}}\label{fig:Rt_S}
\quad\subfloat[Large Area Translation Filter]{\hspace{.37\linewidth}}\label{fig:Rt_L}
\quad\quad\subfloat[Scale Filter]{\hspace{.12\linewidth}}\label{fig:Rs}
\caption{Examples of three filters with hand-crafted features. The desired responses by 
the large and small area translation filters and scale filter are represented by $y_{R_{t}^{L}}$, $y_{R_{t}^{S}}$
and $y_{R_{s}}$.}
\label{fig:Filters}
\end{figure}

\textbf{{\it E}nKCF with Hand-Crafted Features} We, first, use two
conventional hand-crafted features, fHoG (shape)
\cite{felzenszwalb2010object} and color-naming (color)
\cite{li2014scale}. Figure \ref{fig:Filters} shows examples of these
features. This setup of the {\it E}nKCF is designed to perform
tracking at $30$ fps on low-end embedded systems without GPUs.

We use both fHoG \cite{felzenszwalb2010object} and color-naming
\cite{van2009learning} for the large area translation filter,
$R_{t}^{L}$. This is because the fHoG applied to relatively larger
area tends to be more noisy and less discriminative, and adding color
information makes the feature vector for the translation filter more
discriminative. Figure \ref{fig:Filters}(b) shows an example of this
translation filter. By contrast, the $R_{t}^{S}$ only employs the fHoG
features because it covers a relatively smaller area.  Figure
\ref{fig:Filters}(a) shows an example of this translation
filter. Lastly, we use both fHoG and color-naming features again for
the scale filter, $R_{s}$. By assigning fHoG and color-naming
features, we ensure that the likelihood of inaccurate scale estimation
is reduced in comparison to the scale filter with only fHoG
features. This is more important in our case as the scale filter is
operated in every 5 frames in the proposed tracker. Also, the scale
filter, explained earlier in the algorithm \ref{alg:MKCF}, estimates
the scale of a target by correlating it with three candidate
ROIs. This search may increase the run-time complexity of the scale
filter from $\mathcal{O}(n\log n)$ to $\mathcal{O}(3(n\log n))$. To
ensure practical actual filtering operations, we use a smaller
template size (i.e., 64$\times n$) for the scale filter, $R_{s}$. The
smaller template size does not degrade the tracking performance
because of 1) zero padding around the target (smaller ROI) and 2) use
of color-naming features.

\textbf{{\it E}nKCF with Deep Convolutional Features} In addition to
hand-crafted features, we explore deep convolutional features to
extend the applicability of the {\it E}nKCF and boost tracking
performance.

\begin{figure}[!h]
\centering
\includegraphics[width=0.48\textwidth]{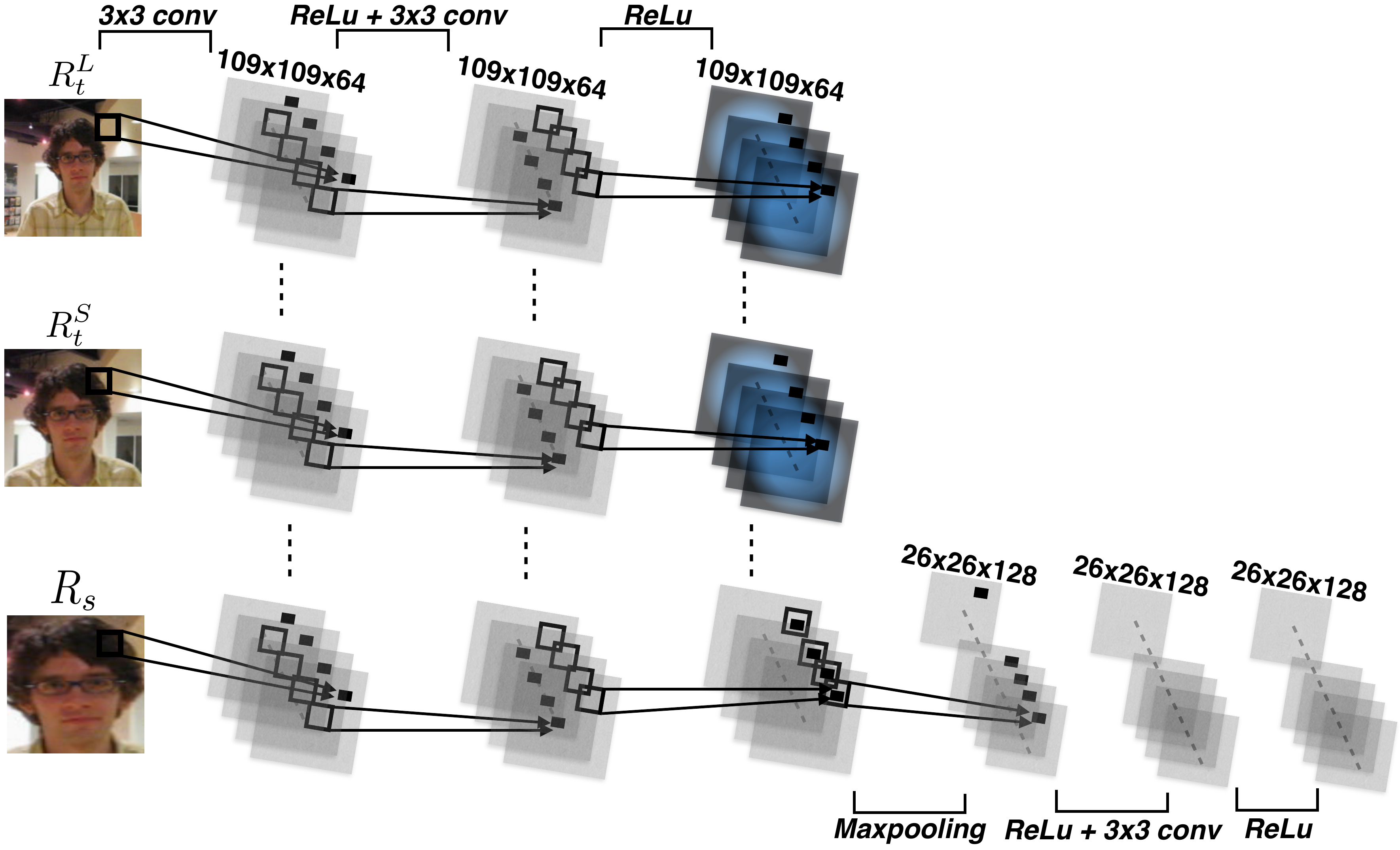}
\caption{The proposed conv224-VGGNet feature extraction in Deep{\it
    E}nKCF. $R_{t}^{L}$ and $R_{t}^{s}$ used an hanning window to
  avoid distortion caused by FFT operation whereas $R_{s}$ does not,
  in order to avoid target boundary information loss.}
\label{fig:Filters_CNN}
\end{figure}

There has been a large volume of studies on utilizing a pre-trained
CNN features in tracking-by-detection algorithms. For example,
\cite{danelljan2015convolutional} used the activation of the first and
second convolutional layer of the VGGNet \cite{simonyan2014very}.
They report that the first convolutional layer features lead to
slightly better precision and success rates than the second layer due
to increased translational invariance in this
layer. \cite{ma2015hierarchical} proposed a KCF tracker integrating
features with higher abstraction capacity.  Their method first employs
the third convolutional layer features to estimate the response
map. In the next stage, the KCF, concentrating on the second
convolutional layer features, is run to update transition. The third
KCF then works on the transition given by the previous KCF and learns
the first convolutional layer features to update the transition. This
coarse-to-fine translation estimation accommodates different levels of
abstractions of the object. However, it runs multiple KCFs in a
sequential fashion, increasing the run-time complexity. For this
reason, we follow an approach similar to
\cite{danelljan2015convolutional}, in order to embed deep features in
     {\it E}nKCF. Our {\it E}nKCF algorithm enables us to exploit
     different level of feature encodings with different KCFs. The
     translation filters, $R_{t}^{L}$ and $R_{t}^{S}$, consider at
     least twice bigger area than the scale filter, $R_{s}$. Given
     this, we can assign deeper feature encodings to $R_{s}$ as it
     learns less spatial information than $R_{t}^{L}$ and
     $R_{t}^{S}$. Thus we assign the activation of the fourth
     convolutional layer features ($26\times26\times128$) to $R_{s}$
     whereas $R_{t}^{L}$ and $R_{t}^{S}$ are assigned the activation
     of the second layer features ($109\times109\times64$). Figure
     \ref{fig:Filters_CNN} illustrates this feature
     assignment. Additionally, we assign the second convolutional
     layer features to $R_{s}$ as in $R_{t}^{L}$ and $R_{t}^{S}$ and
     compare this setting (\textit{conv222-VGG}) to
     \textit{conv224-VGG}. The conv222 setting stands for the
     assignment of the activation of $2^{th}$ convolutional layer
     features of VGGNet to $R_{t}^{L}$, $R_{t}^{S}$ and $R_{s}$. We
     use the VGGNet since it provides higher spatial resolution at the
     first several layers than AlexNet \cite{krizhevsky2012imagenet}.

\textbf{Particle Filter for Smoothing Transition among KCFs} As
explained earlier, the {\it E}nKCF updates the target's scale at every
other $k$ frames. Although the strategy of updating every other $k$th
frames will result in an optimal run-time performance, this may lead
drifts at the later frames. To prevent such potential drifts, we
developed a Bayes filter that incorporates a target's motion to smooth
any intermediate outputs from individual KCFs. In particular, we use a
particle filter to estimate the target's image coordinate based on the
{\it E}nKCF's outputs. The state in this paper, $\boldsymbol{X_{t}}$,
represents the target's pixel coordinates and its velocity,
$\boldsymbol{X_{t}} = \lbrace x, y, v_{x}, v_{y} \rbrace$, where $x$
and $y$ are the pixel coordinates of the target's centroid, $v_x$ and
$v_y$ are the velocities estimated along the $x$-axis and
$y$-axis. The particle filter predicts, using a constant velocity
motion model, the target's next state by generating a predefined
number of particles. Then it uses the confidence maps of {\it E}nKCF
as observation to update its state. The particle weight is computed
as, $w_{p_{t}}(x_{t},y_{t}) = \sum_{i=1}^{N}\sum_{j=1}^{M}
\boldsymbol{y_{R}}(x_{t}-i,y_{t}-j)$, where $w_{p}$ is the weight of
the particle $p$ at time $t$, $\boldsymbol{y_{R}}$ is the response map
from one of the KCFs, and $N$ and $M$ denote the number of rows and
columns of the confidence map.
        
\section{Experiments} \label{sc:Experiments}
So far we explained how the proposed algorithm could efficiently
perform a single-target tracking task while effectively addressing
variations in scale and translation of moving objects. To accurately
evaluate the performance of the proposed algorithm in terms of the
goal of this paper, we need data with challenging variations in scale
and translation. To this end, we chose the UAV123
data\footnote{\url{https://ivul.kaust.edu.sa/Pages/Dataset-UAV123.aspx}}\cite{mueller2016uav123}
that contains 123 video sequences of objects captured from
low-altitude UAVs. Because of the nature of the video acquisition --
the targets and the camera are moving together, this data pose a great
deal of challenges as the scale and position of the targets change drastically
in most of the sequences. To verify that our algorithm is
useful not only for the image data with such extreme challenges but
also for the one with nominal difficulties in single-object tracking,
we also used the OTB100
data\footnote{\url{http://cvlab.hanyang.ac.kr/tracker_benchmark/benchmark_v10.html}}
\cite{WuLimYang13}
that contains the videos recorded from smart phones and typical
cameras in perspective view. In addition, we use the temporarily
down-sampled UAV123 dataset to check how robust our algorithm is to
drastic motions of camera and moving targets. Lastly, we evaluate the
proposed algorithm with deep convolutional features on the UAV123
dataset to investigate how much performance gain we can achieve in
using deep features over the conventional features like fHoG and
color-naming.

\textbf{Finding Optimal Hyper-parameters} As each of three KCFs in
       {\it E}nKCF is designed to address specific challenges in
       single target tracking problem, the optimal parameters for
       individual KCFs should be different. We set the learning rates ($\beta$)
       of individual filters, $R_{t}^{L}$, $R_{t}^{S}$, and $R_{s}$ as
       $0.020$, $0.020$ and $0.010$. For the kernel of the Gaussian
       neighboring function, we empirically found the optimal values
       of $\alpha$ as $0.7$, $0.6$, and $0.9$ for $R_{t}^{L}$,
       $R_{t}^{S}$, and $R_{s}$. We set the scale filter update
       threshold, $T_{R_{s}}$, to 4, peak-to-sidelobe (PSR) ratio. The
       padding size for the correlation filters is tuned to $2$ for
       $R_{t}^{L}$, $1.50$ for $R_{t}^{S}$, and $0$ for $R_{s}$. For
       our particle filter implementation, we empirically set the
       number of particles to $1,000$ to balance the run-time
       performance and accuracy. To keep the level of variance among
       the particles reasonable, we performed the re-sampling only
       when the \textit{efficient number of samples}, ($ \hat{N}_{eff}
       \approx (\sum_{p=1}^{P}\boldsymbol{w}_{p}^{2})^{-1} $), is lower
       than a pre-defined threshold.

\begin{figure*}[h]
\centering
\begin{tabular}{ccc}
\tiny\textbf{UAV123}\hspace{.43\linewidth}\textbf{OTB100}\\
\includegraphics[width=3.80cm]{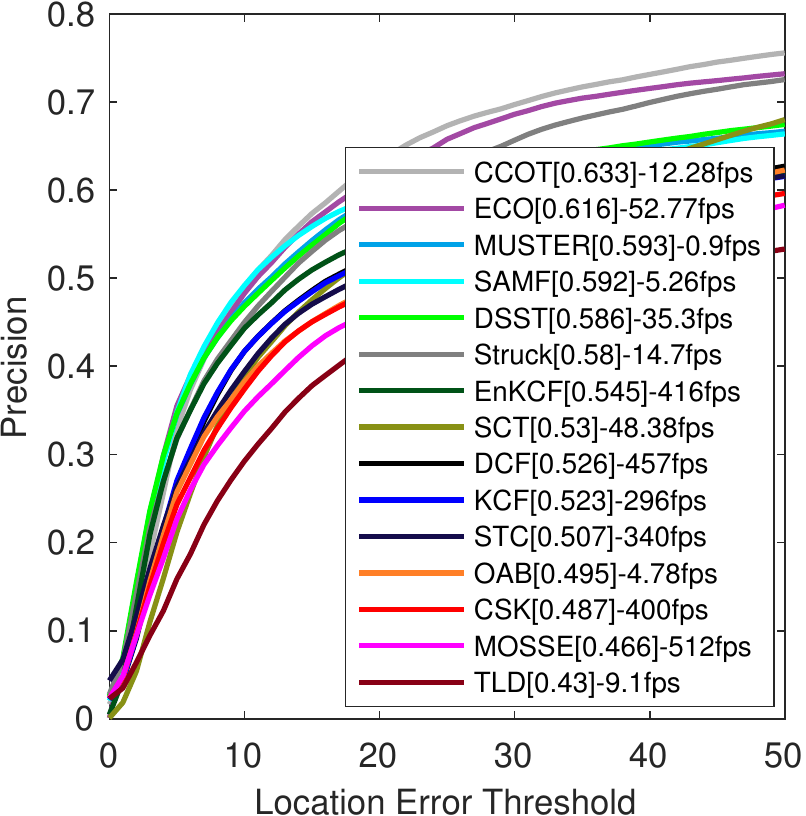}
\includegraphics[width=4.10cm]{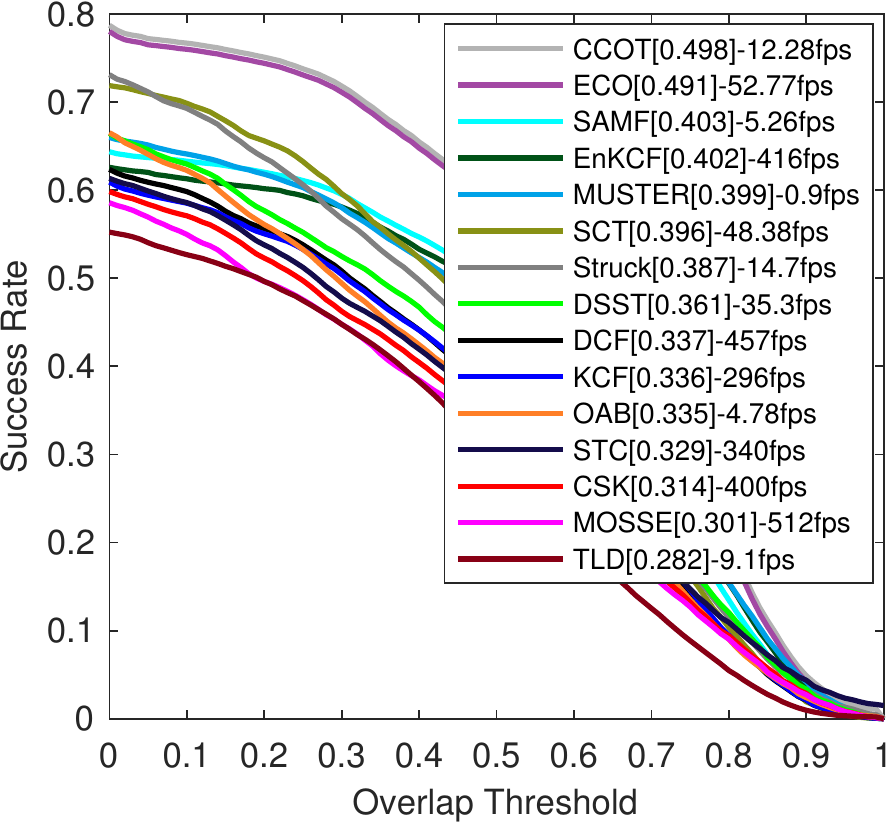}
\includegraphics[width=3.80cm]{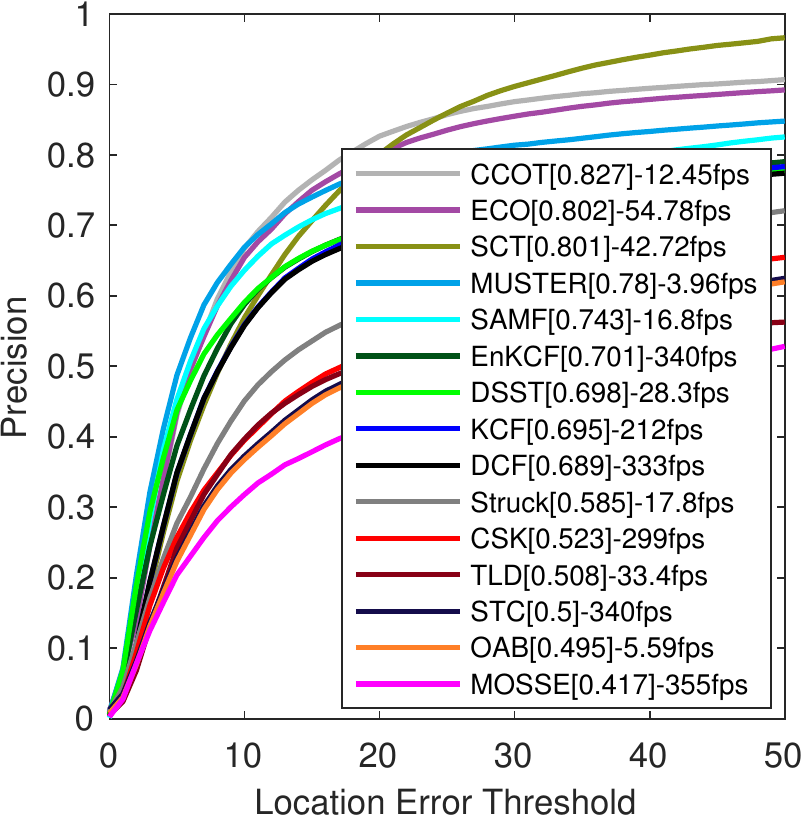}
\includegraphics[width=4.10cm]{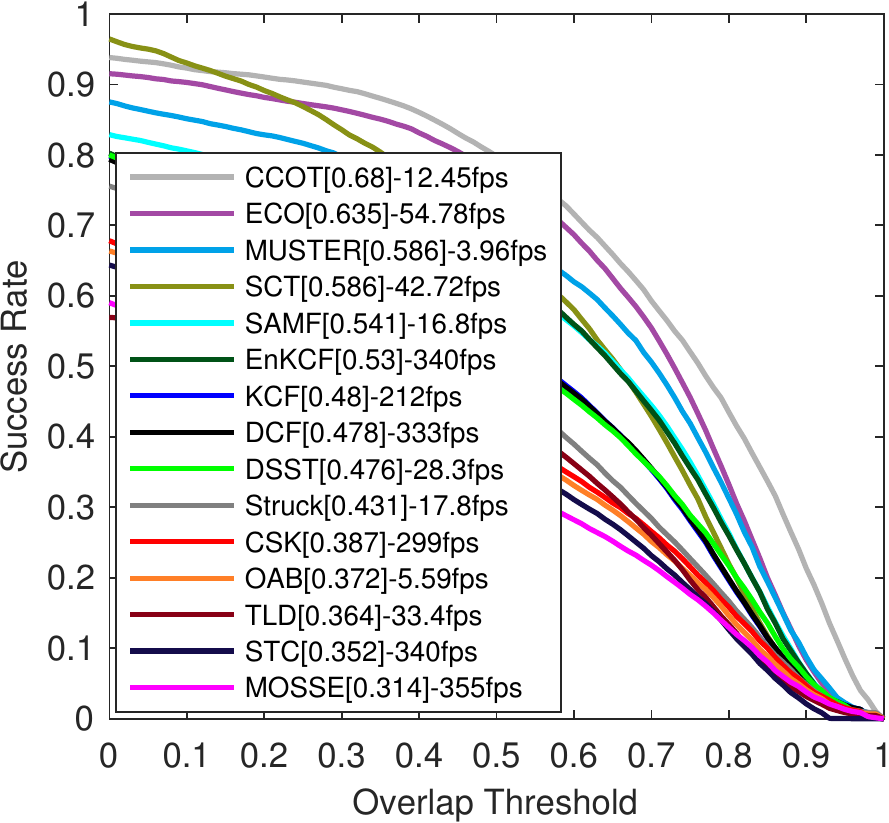}\\
\end{tabular}
\caption{Comparison of {\it E}nKCF's performance with other trackers on the UAV123 and OTB100 datasets.}
\label{fig:UAV123_DATASET}
\end{figure*}

\textbf{Performance on UAV123 Dataset} We used the precision and
success rates to compare the performance of the proposed algorithm
with those of the state-of-the-art tracking algorithms. For the
precision, we rank the trackers based on the precision numbers at 20
pixels whereas in the success rate plots, they are ranked based on the
area under curve (AUC) scores. The tracking algorithms under the
comparison include ones in high-speed ($\geq$300 fps): KCF
\cite{henriques2015high}, CSK \cite{henriques2012exploiting}, DCF
\cite{henriques2015high}, MOSSE \cite{bolme2010visual}, and STC
\cite{zhang2014fast} and ones in relatively lower-speed ($\leq$50):
ECO \cite{DanelljanCVPR2017}, CCOT \cite{DanelljanECCV2016}, SCT
\cite{Choi_2016_CVPR}, SAMF \cite{li2014scale}, DSST
\cite{danelljan2014accurate}, Struck \cite{hare2012efficient}, MUSTER
\cite{hong2015multi}, TLD \cite{kalal2012tracking}, and OAB
\cite{zhang2012robust}. The ECO and CCOT trackers originally uses deep
convolutional features, however, to perform fair comparison, we employ
fHoG and color-naming features similar to our tracker. Figure
\ref{fig:UAV123_DATASET} shows the results on the UAV123 dataset. The
    {\it E}nKCF outperformed other high-speed trackers by $3\%$-$15\%$
    at 20 pixels precision. In particular, three algorithms, SAMF,
    DSST, and Struck did about $5\%$ better than ours in accuracy, but
    10-20 times slower. The more recent trackers, ECO and CCOT, on the
    other hand, outperforms {\it E}nKCF by $15\%$ while running at
    10-15 times slower. For the scale adaptation, {\it E}nKCF did
    fourth best in terms of AUC for the success rate plot. It
    outperformed other high-speed trackers by about $20\%$-$25\%$ in
    AUC. In addition, it performed even better than some of the
    lower-speed trackers.  For example, for AUC, {\it E}nKCF
    outperformed Struck and DSST by $5\%$ and $10\%$ while running at
    more than $10$ and $30$ times faster. For the DSST, we believe our
    algorithm outperformed it because, first of all, the DSST uses a
    1-D scale filter and searches the target's scale over 33
    candidates in a scale pool whereas our algorithm uses a 2-D scale
    filter with 3 candidate scales. Learning a 2-D filter resulted in
    learning more spatial information. Second of all, the DSST uses
    only fHoG whereas our algorithm uses both fHoG and color-naming
    features. Learning complementary features such as color and shape provided a 
    better understanding of the ROI.
    
\begin{table*}[!h]
\smaller
\begin{center}
\color{green}--- Best \hspace{.15\linewidth}\color{red}--- $2^{nd}$ Best \hspace{.15\linewidth}\color{blue}--- $3^{th}$ Best \color{black}\\
\resizebox{\textwidth}{!}{\begin{tabular}{|l|c|c|c|c|c|c|c|c|c|}
\hline
Method & {\it E}nKCF & $R_{t}^{S}$+$R_{t}^{L}$+$R_{s}$ & $R_{t}^{L}$+$R_{s}$ & $R_{t}^{S}$+$R_{s}$ & $R_{t}^{L}$* & $R_{t}^{S}$* & $R_{t}^{L}$+$R_{s}$* & $R_{t}^{S}$+$R_{s}$* & $R_{t}^{L}+R_{t}^{S}$+$R_{s}$* \\
\hline\hline
Pr. ($20$px) & \color{blue}53.9 & 48.93 &52.41 & 48.10 & 51.88 & 51.29 & \color{red} 55.85 & 52.14 & \color{green}58.16 \\
SR ($50\%$) & \color{red} 40.2 & 36.75 & 38.23 & 36.04 & 35.12 & 34.43 & \color{blue}39.89 & 38.51 & \color{green}41.58 \\
FPS & \color{red} 416 & \color{blue}412 & 370 & \color{green}425 & 365 & 384 & 135 & 151 & 99 \\
\hline
\end{tabular}}
\end{center}
\caption{Results of running different combinations of the KCFs for
  UAV123 dataset. The '$*$' represents a sequential approach where
  multiple KCFs are, in the given order, applied to every frame like
  in LCT \cite{ma2015long}.}
\label{table:Comparison_to_LCT}
\end{table*}

\textbf{Finding Optimal Combination and Deployment Order of Multiple
  KCFs} In addition to performance comparison with other trackers, we
also conducted an experiment to see what combination and deployment
order of the KCF works best. Table \ref{table:Comparison_to_LCT}
presents results of this experiment. To achieve fair comparison, we did
not use the particle filter. The combination presented at the $2^{nd}$
column switched the order of deploying $R_{t}^{L}$ and
$R_{t}^{S}$. The one at the $3^{th}$ column removed $R_{t}^{S}$ and
replaces with $R_{t}^{L}$ whereas the one at the $4^{th}$ column
removes $R_{t}^{L}$. The combinations with the ``*'' marker ran the
filters on every frame in the listed order. The trackers at the
$7^{th}$ and $8^{th}$ columns ran a translation and a scale filter on
every frame which is similar to the LCT tracker
\cite{ma2015long}.\footnote{To be precise, the LCT tracker comes with
  a re-detection module for a long-term tracking.} This experiment
empirically verified that \textit{E}nKCF is an optimal way of using
multiple KCFs, in terms of both tracking accuracy and run-time
performance. For example, one could achieve $4.26\%$ higher in accuracy
by running $R_{t}^{L}$ and $R_{S}$ at every frame, but this
combination decreases run-time performance to 317 fps. 

\textbf{Evaluation on Particle Filter Contribution} For the UAV123
dataset, we also evaluated the performance of the {\it E}nKCF with and
without the particle filter. In particular, we use 50 video sequences
with no drastic camera motion. To evaluate the robustness of particle
filter, we add noise from a uniform distribution ($[-20,20]$), to the
translation estimations of the KCFs. For this experiment, the {\it
  E}nKCF with particle filter achieves $51.98\%$ precision ($20$px),
outperforming the one without particle filter by $6.32\%$. This
experiment validated that the integration of particle filter into the
{\it E}nKCF can further improve the performance.

\textbf{Performance on OTB100 Dataset} Figure \ref{fig:UAV123_DATASET}
shows the results on the OTB100 dataset. The performance of {\it
  E}nKCF on the OTB100 data is similar to that of the UAV123
dataset. Specifically, it performed reasonably well at estimating
target scale in that it showed the highest precision and success rates
among the other high-speed trackers. Interestingly, it outperformed
another correlation filter based tracker, DSST, but performing $5\%$
behind of another low-speed scale adaptive SAMF tracker. Finally, the
ECO, and CCOT achieve $15\%$-$20\%$ higher precision and success rates
than ours while operating at 10-20 times slower rates.

\begin{figure}[!h]
\centering
\begin{tabular}{cc}
\tiny\quad\textbf{UAV123$\_$10fps}\\
\includegraphics[width=4.02cm]{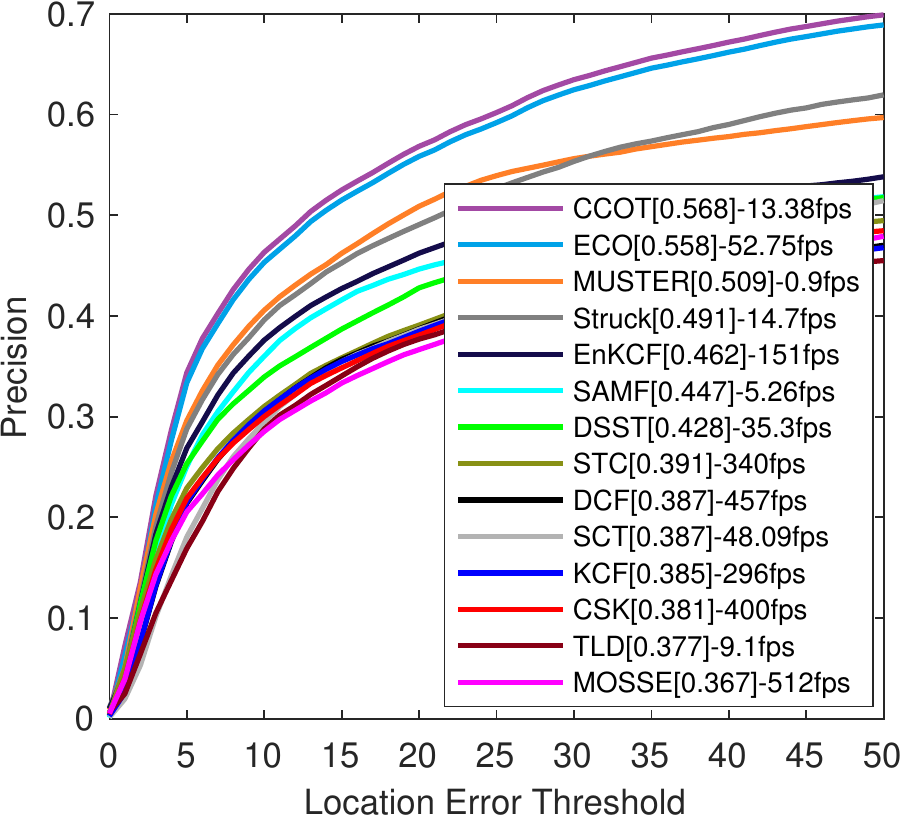}
\includegraphics[width=4.00cm]{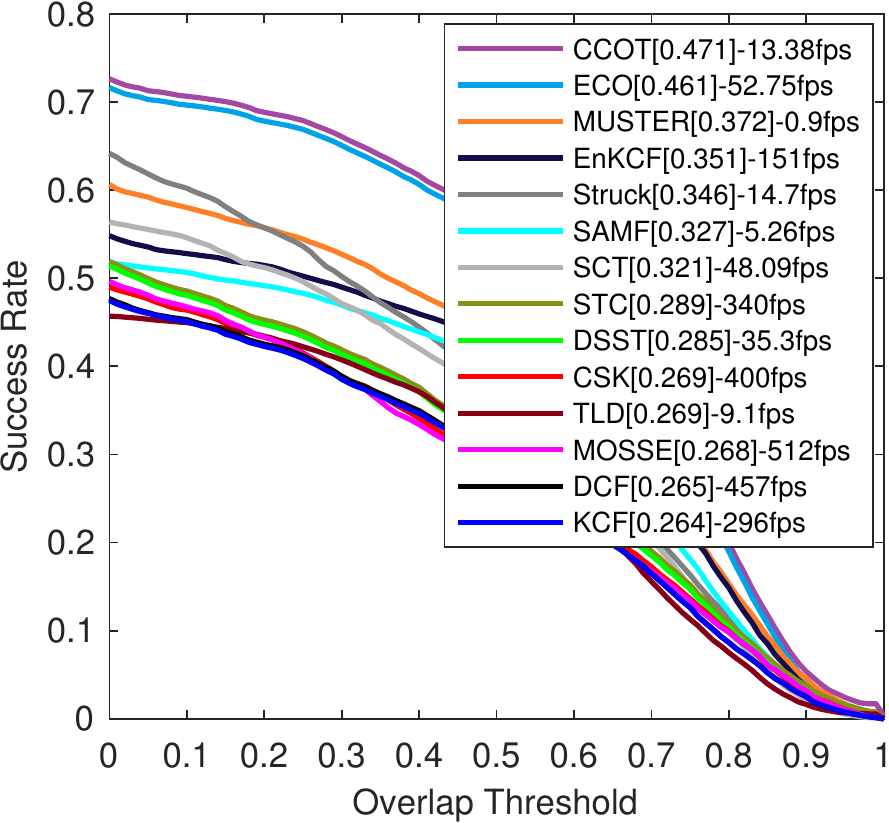}\\
\end{tabular}
\caption{Comparison of a modified {\it E}nKCF's performance with those
  of the state-of-the-art tracking algorithms on a lower frame rate
  data.}
\label{fig:UAV123_10fpsDATASET}
\end{figure}

\textbf{Performance on UAV123$\_$10fps Dataset} We were curious about
how the frame rate of a testing video would affect the performance of
a tracking algorithm. To this end, we use the temporarily down-sampled
version of the UAV123, called UAV123$\_$10fps dataset. We believe the
downsampling would make the original UAV123 data more challenging
because the displacements of the moving objects become even bigger. To
tackle this challenge, we slightly modified the proposed algorithm --
ran $R_{t}^{L}$ every frame and removed the particle filter due to
larger ego-motion. Figure \ref{fig:UAV123_10fpsDATASET} shows the
performance of the modified version of {\it E}nKCF and other tracking
methods. The precision rates of the ECO and CCOT dropped about $10\%$
in comparison to the original UAV123 dataset. Our tracker outperforms
other high-speed correlation filter based trackers including KCF, DSST
by about $15\%$-$20\%$ and showed a precision rate similar to that of
SAMF. It ranks as fourth in AUC while running about $10$ to $100$
times faster than the top three trackers. It was interesting to
observe how sensitive the performance of the state-of-the-art tracking
algorithms is to the frame rate, and, with a slight modification, the
proposed method effectively handled large displacement of objects in
successive frames.

\textbf{Evaluation with Deep Features} Our focus so far was to develop
a single-target tracker that does not require offline learning to
effectively address the variations in scale and translation, and can
run at a high-speed, $\ge 300$fps. We would like to see how much the
performance gain we could achieve by replacing the conventional
features, fHoG and color-naming with deep convolutional features. We
use deep convolutional features similar to \cite{ma2015hierarchical,
  danelljan2015convolutional}. Note that running the {\it E}nKCF with
deep features would obviously increase the computational cost and
degrade the run-time performance. 

\begin{figure}[!h]
\centering
\begin{tabular}{ccc}
\tiny\quad\quad\textbf{UAV123}\\
\includegraphics[width=3.60cm]{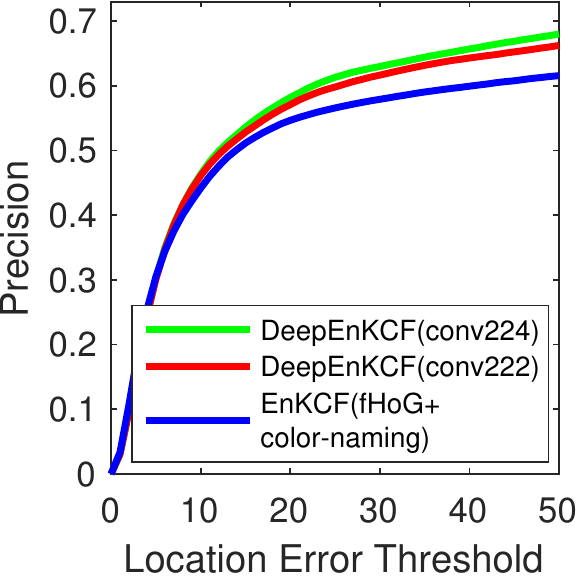}
\includegraphics[width=3.70cm]{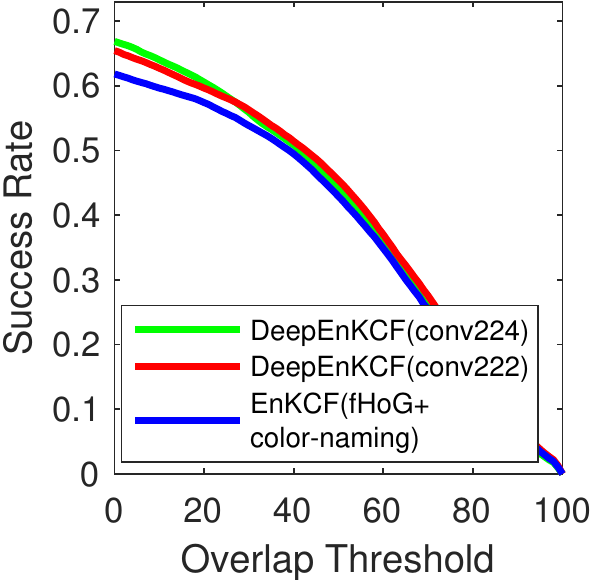}\\
\end{tabular}
\caption{Comparison of {\it E}nKCF's ({\it fHoG} + {\it color-naming})
  performance with Deep{\it E}nKCF on the UAV123 dataset. The frame
  rate of the Deep{\it E}nKCF (conv224) was, on average, 30.74 fps and
  the conv222 setting was 35.23 fps on a CPU.}
\label{fig:UAV123_DATASET_DeepFeatures}
\end{figure}

Figure \ref{fig:UAV123_DATASET_DeepFeatures} shows the DeepE{\it
  n}KCF's performance on UAV123 dataset. Deep{\it E}nKCF outperformed
the {\it E}nKCF with hand-crafted features by about $3\%$ to $5\%$ in
precision (20 px) and $2\%$ in success rates (AUC). The
\textit{conv224-VGG} setting performed slightly better than
\textit{conv222-VGG} in precision while achieving similar success rate
(AUC). This indicates that higher level feature abstraction works
better for the smaller ROIs. By using feature abstractions from the
VGGNet pre-trained on millions of images, we can better represent the
low level features of the object than hand-crafted features in
challenging cases such as low contrast ROIs. However, we believe that
the contribution of the deep features is limited by two factors:
increased translational invariance in deeper layers and losing targets
due to large target and camera motion in the UAV123 dataset. 

\section{Conclusion} \label{sc:Conclusion}
Running a computer vision algorithm on any existing embedded systems
for real-world applications is economically and practically very
attractive. Among other practical considerations, it would be
desirable if such computer vision algorithms require no offline
training and operate at high-speed. To develop a single-target
tracking algorithm to meet these features, we proposed an extension of
KCF that applies three KCFs, in turn, to address the variations in
scale and translation of moving objects. We also developed a particle
filter to smooth the transition between three KCFs, especially the
transition between the scale and large ROI translation filter. Through
experiments, we found that the way the {\it E}nKCF deployed three KCFs
was optimal, and the particle filter contributed to increase the
performance.  We used two public datasets to evaluate the performance
of the proposed algorithm and other state-of-the-art tracking
algorithms, and found that, on average, the performance of the
proposed algorithm is better than other high-speed trackers over 5\%
on precision at 20 pixels and 10-20\% on AUC. Our implementation ran
at 340 fps for OTB100 and at 416 fps for UAV123 data that is faster
than DCF (292 fps) for OTB100 and KCF (292 fps) for UAV123. Finally,
we explored the idea of utilizing deep features for the proposed
algorithm and found that the deep features helped the proposed
algorithm boost the performance by $10\%$.

Although the proposed algorithm showed a promising result tested with
challenging data, we believe the corner case analysis has not been
extensively done yet. As future work, we would like to thoroughly
study under what conditions our algorithm would fail to track 
objects. In addition, we also would like to investigate a way
of re-initializing a target when the target is lost, due to occlusion,
illumination change, drastic camera motion, etc.

{\small
\bibliographystyle{ieee}
\bibliography{draft}
}

\end{document}